\documentclass[journal]{IEEEtran}
\usepackage{cite}
\usepackage{amsmath,amssymb,amsfonts}
\usepackage{color}
\usepackage{graphicx}
\usepackage{textcomp}
\usepackage{caption}
\usepackage{verbatim}
\usepackage{amssymb}
\usepackage{booktabs}
\usepackage{multirow}
\usepackage{verbatim}
\usepackage{enumerate}
\usepackage{graphicx}
\usepackage{subfig}
\usepackage{algorithm}
\usepackage{algorithmicx}
\usepackage{wasysym}
\usepackage{amssymb}
\usepackage{amsmath, xparse}
\usepackage{hyperref}
\usepackage{array}
\usepackage[noend]{algpseudocode}
\algrenewcommand\algorithmicrequire{\textbf{Input:}}
\algrenewcommand\algorithmicensure{\textbf{Output:}}

\newcommand{\qin}[1]{{\textcolor{green}{#1}}}

\begin{document}
\title{Multilevel Saliency-Guided Self-Supervised Learning for Image Anomaly Detection}
\author{Jianjian Qin$^{*}$,  Chunzhi Gu$^{*}$, Jun Yu, Chao Zhang
\thanks{$^{*}$These two authors contributed equally to this work.}
\thanks{C. Zhang (zhang@u-fukui.ac.jp) and J. Qin are with the University of Fukui, Japan.}
\thanks{C. Gu is with the Toyohashi University of Technology, Japan.}
\thanks{J. Yu is with the Niigata University, Japan.}}

\maketitle

\begin{abstract}
 Anomaly detection (AD) is a fundamental task in computer vision. It aims to identify incorrect image data patterns which deviate from the normal ones. Conventional methods generally address AD  by preparing augmented negative samples  
to enforce self-supervised learning. However, these techniques typically do not consider semantics during augmentation, leading to the generation of unrealistic or invalid negative samples. Consequently, the feature extraction network can be hindered from embedding critical features. In this study, inspired by visual attention learning approaches, we propose CutSwap, which leverages saliency guidance to incorporate semantic cues for augmentation. Specifically, we first employ LayerCAM to extract multilevel image features as saliency maps and then perform clustering to obtain multiple centroids. To fully exploit saliency guidance, on each map, we select a pixel pair from the cluster with the highest centroid saliency to form a patch pair. Such a patch pair includes highly similar context information with dense semantic correlations. The resulting negative sample is created by swapping the locations of the patch pair. Compared to prior augmentation methods, CutSwap generates more subtle yet realistic negative samples to facilitate quality feature learning.
Extensive experimental and ablative evaluations demonstrate that our method achieves state-of-the-art AD performance on two mainstream AD benchmark datasets.


\end{abstract}

\begin{IEEEkeywords}
Anomaly detection, Data augmentation, Self-supervised learning, Representation learning
\end{IEEEkeywords}

\section{Introduction}
\label{sec:introduction}
Industrial anomaly detection (AD) aims to distinguish defective samples that cannot satisfy the real-world requirements in industry. It has been actively studied in recent years because of its wide range of applications such as manufacturing detection \cite{wang2021student_teacher, de2022hybrid, ko2022new, qin2023teacher,jang2023n,wu2023physics} and biomedical analysis \cite{li2023self, cho2023training, pinaya2022unsupervised,wolleb2022diffusion,sivapalan2022annet,fernando2020neural}. The key issue of AD is that, because acquiring a sufficient number of anomalous samples is virtually impractical, data augmentation is often required prior to applying machine learning detection algorithms \cite{bozorgtabar2023attention, schluter2022natural, long2022self}.

\begin{figure}[t]
\centering
\subfloat[Normal] {\includegraphics[width=25mm,scale=0.5]{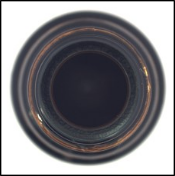}}\hspace{1.5em}%
\subfloat[Anomalous] 
{\includegraphics[width=25mm,scale=0.5]{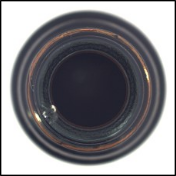}}\hspace{1.5em}%
\\[0.2ex]

\subfloat[CutPaste (B2B)] {\includegraphics[width=25mm,scale=0.5]{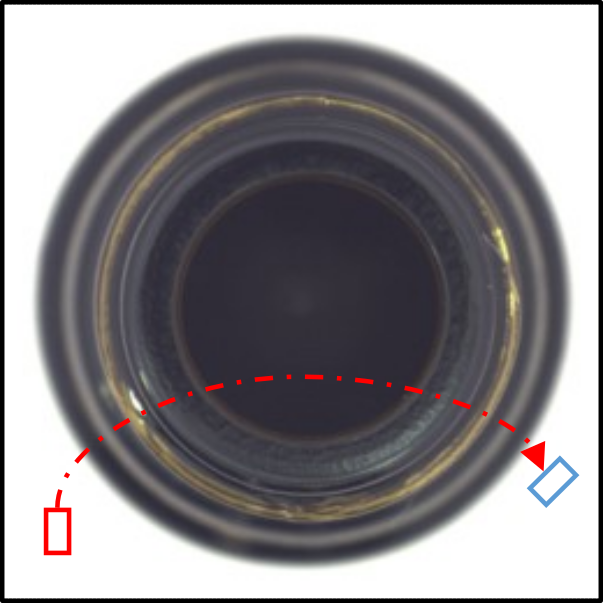}}\hspace{0.5em}%
\subfloat[CutPaste (B2F)] 
{\includegraphics[width=25mm,scale=0.5]{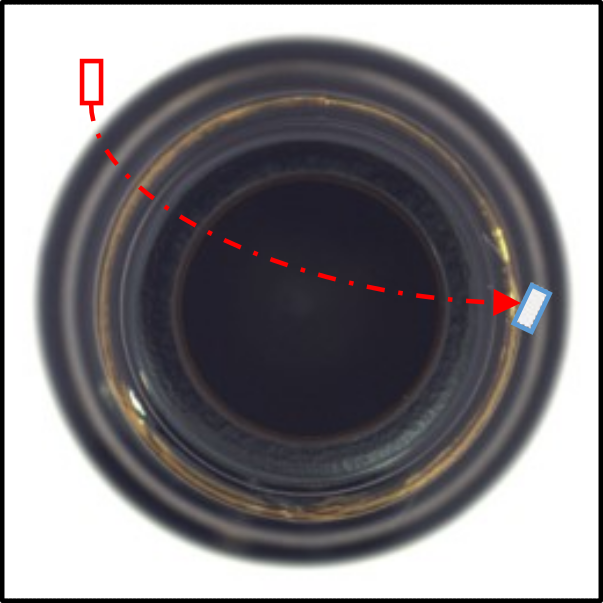}}\hspace{0.5em}%
\subfloat[CutSwap] 
{\includegraphics[width=25mm,scale=0.5]{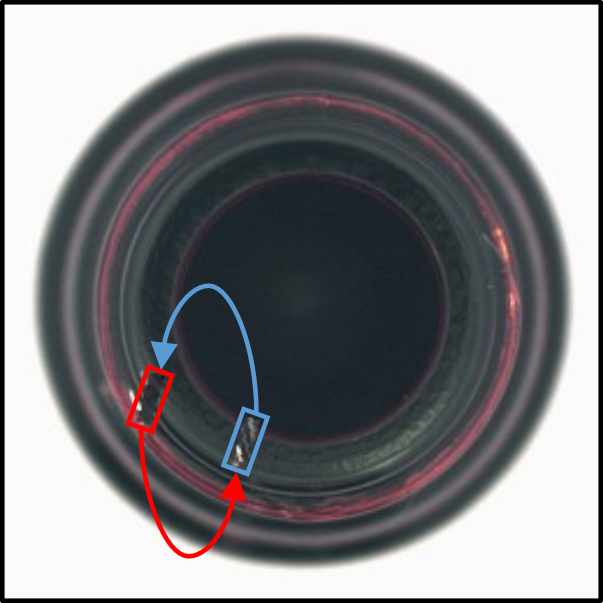}}\hspace{0.5em}%
\caption{\textbf{An example of conceptual comparison of CutSwap (our approach) against CutPaste \cite{li2021cutpaste}.}  Given a source anomaly-free sample (a), \cite{li2021cutpaste} can easily induce less realistic augmentation results (e.g.,  background-to-background (B2B) exchange in (c)), or unnatural anomalies (e.g., foreground-to-background (B2F) exchange in (d)). Our proposed CutSwap augmentation yields subtle yet realistic anomalies which appear similar to the true anomalous sample (b).}
\label{fig:1}
\end{figure}

Earlier data augmentation includes trivial operations, such as blurring \cite{hussain2017differential,shorten2019survey}, noising \cite{bae2018perlin,moreno2018forward}, or rotating \cite{xi2018sr,liu2020task} the existing anomaly-free samples. Although these 
techniques can conveniently produce a large number of additional anomalous samples, the synthetic anomalous patterns tend to bias from the real patterns, leading to low generality of the models. Thus, data augmentation approaches have shifted towards more dedicated augmenting policies \cite{liang2023miamix,trabucco2023effective,hao2023mixgen,xu2023comprehensive,liu2023anomaly}. Among these, a recent method, CutPaste \cite{li2021cutpaste}, offers a novel concept of augmentation. It first randomly cuts a patch from the positive\footnote{A positive sample refers to a sample obtained via applying weak augmentation techniques (e.g., color jittering) to a normal sample.} image sample and then pastes it back on a random position of this image again to form a negative sample which includes subtle defects. By performing self-supervised learning with the resulting positive-negative sample pair to construct a feature extractor, downstream tasks (e.g., AD) can be effectively handled. Despite the high anomaly fidelity compared with previous augmentation schemes, CutPaste can also easily result in undesired negative samples. As shown in Fig. \ref{fig:1}(c), once the cutting and pasting positions are both located on the background (B2B), the produced ``anomalous'' sample would be identical to the original normal one. Similarly, if the background patch is pasted on the foreground object area, the anomalous sample would (B2F, Fig. \ref{fig:1}(d)), in contrast, seems highly unrealistic. Consequently, these wrongly augmented pairs can significantly mislead the 
feature extractor from learning quality latent representations for AD. Although a later work \cite{zou2022spot} improves this by introducing spot areas for pasting to ensure anomaly realism, the potential disadvantage imposed by random pasting remains. Furthermore, the generation of undesired samples in these methods stems from the lack of semantic awareness in augmentation.

In this study, we propose a novel saliency-guided augmentation strategy to facilitate self-supervised representation learning for AD. Following the augmentation policy of CutSwap, our core idea is to leverage the saliency map to provide augmentation guidance such that the cutting position focuses centrally on an important image region. Specifically, we first leverage LayerCAM \cite{jiang2021layercam} as a saliency extractor to obtain latent representations from normal samples. LayerCAM captures the coarse-to-fine CNN activation to reflect multilevel saliency maps. These maps collect object localization information from rough- to fine-grained levels to highlight object-related pixels, which can be regarded as anchors to guide augmentation. We then directly perform clustering on the produced saliency intensity to localize the anchor position. For the produced cluster centroid with the maximum saliency intensity, we next randomly select two pixels from this cluster as anchors to generate a patch pair. Ideally, the information inside the two patches should be semantically close with respect to the regions of high-interest confidence. Motivated by this, we swap the location of the patch pair to create the negative sample. Instead of pasting, which can potentially damage semantics, the swap operation yields a more challenging anomaly setting and attempts to best retain the original semantics. 
Thus, we term our method as CutSwap, which is a semantic-aware strategy for negative sample augmentation. As illustrated in Fig. \ref{fig:1}(e), CutSwap synthesizes more subtle yet realistic anomalies. Because LayerCAM outputs multilevel maps, for every positive sample, we prepare multiple paired data by creating a negative sample from some predetermined saliency layer indices. 

Since our goal is to provide a solution to image AD, we follow \cite{li2021cutpaste} by enforcing self-supervised learning to train a feature extraction network on the CutSwap augmented sample pair to embed quality feature representations. During testing, we exploit the Patch Core \cite{roth2022towards} to derive the final anomaly score. We conduct experiments on two large-scale anomaly detection datasets: MVTec AD \cite{bergmann2021mvtec} and VisA \cite{zou2022spot}. The results demonstrate that CutSwap equipped with PatchCore achieves state-of-the-art AD performance compared with prior methods.

The main contributions of this study are as follows:
\begin{itemize}
\item We propose CutSwap, a multilevel saliency-guided data augmentation method by cutting and swapping from high-saliency regions to produce coarse-to-fine levels of negative samples.

\item We introduce a clustering procedure in CutSwap to efficiently and effectively select the target pixels from the saliency map to encourage semantic-aware augmentation.

\item We solve the task of image AD by performing self-supervised learning on the CutSwap augmented sample pair, and report extensive experimental results to demonstrate the effectiveness of our method against competitors.

\end{itemize}

\section{Related work}

In this section, we first introduce the current mainstream visual anomaly detection methods. We then discuss some data augmentation approaches for AD. Finally, we review saliency-guided methods for representation learning.

\subsection{Visual anomaly detection}
The mainstream visual anomaly detection methods \cite{gudovskiy2022cflow, bergmann2020uninformed, roth2022towards, wang2021student,liu2023deep, liu2023simplenet,xie2023iad} can be categorized into unsupervised \cite{tien2023revisiting,li2021cutpaste,lei2023pyramidflow,zou2022spot,madan2023self, yan2021learning, yao2023focus,zhang2023unsupervised, hotta2023subspace} and supervised \cite{sohn2023anomaly, zhang2023prototypical, venkataramanan2020attention,liznerski2020explainable} approaches. Unsupervised AD can be further classified into feature embedding-based and reconstruction-based methods. Feature embedding-based methods \cite{deng2022anomaly, cao2023anomaly, roth2022towards} typically first learn a powerful feature extractor to embed quality latent representations and then perform clustering or density estimation to calculate the distance between each sample and the normal ones in the feature space.  A representative work of the feature embedding-based approach is Patch Core \cite{roth2022towards}, which constructs a memory bank to store patch-level features to distinguish normal and anomalous samples using nearest neighbor calculation. In terms of reconstruction-based methods \cite{schluter2022natural,ristea2022self,tao2022unsupervised}, they attempt to measure the reconstruction error with generative models (i.e., Auto Encoder) to identify the abnormality of unseen samples for AD. For example, Tao et al., \cite{tao2022unsupervised} equipped the reconstruction process with inpainting learning via a dual-siamese framework to enable learning semantic representations. However, as pointed out in \cite{qin2023teacher}, reconstruction-based methods 
are prone to sacrificing fine-level details to yield decent global reconstruction capability. Compared with unsupervised methods, some recent techniques resolve AD by directly providing ground-truth anomaly supervision. Zhang et al., \cite{zhang2023prototypical} formulated Prototypical Residual Network (PRN) to 
 supervise the anomaly segmentation learning such that varying scales of anomaly patterns can be well characterized. Nevertheless, supervised AD has, to date, been less explored because collecting anomalous data covering all possible anomaly types is less practical.\par

\subsection{Data augmentation for AD}
Recently, self-supervised AD techniques \cite{zhao2021anomaly} have been actively studied to encourage the deep nerual network to extract better feature embeddings. Self-supervised AD, which does not have access to anomalous samples, performs data augmentation to manually increase data patterns. In general, the augmented normal and anomalous samples are termed positive and negative samples, respectively, which are leveraged to guide network learning. Therefore, a powerful augmentation method plays a key role in the self-supervised AD. A pioneering work $-$ SimCLR \cite{chen2020simple} introduced random cropping and color distortion to impose global editing on the original images to form positive sample pairs. 
However, the augmented images obtained via SimCLR appear to be less realistic than the real images. Therefore, later efforts \cite{cai2023dual, li2021cutpaste, zou2022spot, huang2022self} focused primarily on encouraging sample realism. CutPaste \cite{li2021cutpaste} randomly crops and then pastes a small patch onto the original image to produce a visually subtle negative sample. Zou et al., \cite{zou2022spot} argued that a triangle patch can still cause pixel inconsistency and instead adopted spot-shaped cutting areas to improve CutPaste \cite{li2021cutpaste}. Huang et al., \cite{huang2022self} added masked regions to the input normal image as anomalous regions and aimed to predict the mask via a reconstruction process. Despite their satisfactory detection accuracy, these methods do not consider semantics and simply sample random pixel positions during augmentation. In contrast, our CutSwap augmentation fully utilizes saliency semantics to achieve natural yet subtle negative samples.

\begin{figure*}[tb]
 \centering
 \includegraphics[width=1\textwidth]{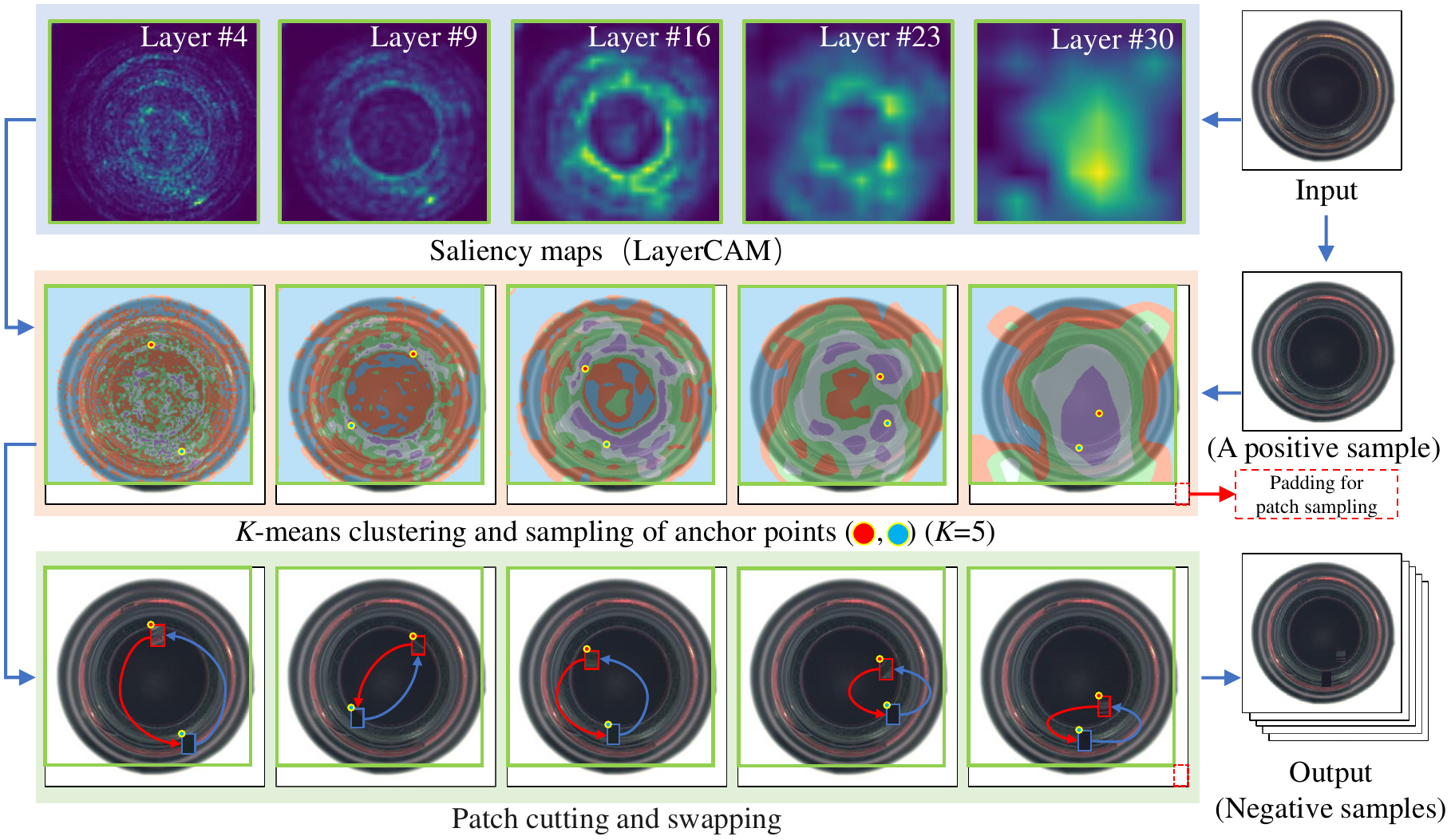}
 \caption{\textbf{Overview of CutSwap augmentation.} First, a normal sample is fed into LayerCAM to extract multi-scale saliency maps (1st row). The corresponding positive sample is also obtained via applying weak augmentation to the normal sample. Then, we apply $K$-means on each saliency map for clustering the saliency score (2nd row). Next, we randomly select two anchor points 
 within the cluster whose centroid has the largest saliency score to create a patch pair. Finally, we cut and then swap the patch pairs to form negative samples.}
 \label{fig:2}
 \end{figure*}

\subsection{Saliency-guided method for representation learning}
There are also some recent studies  \cite{liu2022sisl, chen2023saliency,jiang2021saliency,miangoleh2023realistic} that incorporate saliency in augmentation for representation learning. These methods generally make use of the most salient regions in the image and treat the saliency map as an alternative to the segmentation map. Liu et al., \cite{liu2022sisl} regarded saliency score as segmentation guidance and cropped salient regions to create positive samples for image classification. 
A similar augmentation strategy is developed in 
 \cite{chen2023saliency} for object detection on scene images. Jiang et al., \cite{jiang2021saliency} designed a shift mask according to the saliency map to realize versatile image editing tasks. Miangoleh et al.,  \cite{miangoleh2023realistic} devised a saliency loss to penalize unrealistic image edits. In contrast to these methods, our approach resorts to multiscale saliency guidance to best interpret visual attention at different granularity for augmentation for the task of AD.

\section{Method}
\label{Method}
Let us now introduce our approach $-$ CutSwap, to self-supervised learning for image AD. The goal of CutSwap is to augment quality anomalous samples such that the representation learning network can be well empowered to extract quality deep features even for unseen samples. In general, CutSwap resorts to saliency guidance to enable the created negative sample to be semantically coherent with a positive sample. Formally, by assuming the original normal data set to be $\mathcal{X}$, a weak augmentation procedure (e.g., color jittering or rescaling) is first introduced to produce a positive sample set 
$\widetilde{\mathcal{X}}^{+}$. CutSwap augmentation aims to generate a quality negative sample set $\widetilde{\mathcal{X}}^{-}$ for pairing $\widetilde{\mathcal{X}}^{+}$ to realize self-supervised learning. In the following sections, we first detail the CutSwap augmentation in Sec. \ref{sec:Cutswap}. We then describe the anomaly detection pipeline using our self-supervised framework in Sec. 
\ref{sec:self-supervised learning framework}. \par

\subsection{CutSwap}
\label{sec:Cutswap}
Here, we introduce the CutSwap augmentation to yield $\widetilde{\mathcal{X}}^{-}$. In contrast to the CutPaste augmentation that can potentially cause less realistic samples, we aim to devise a more elegant augmentation manner that naturally provides semantic clues. To this end, motivated by the effectiveness of attention-based localization techniques \cite{chen2023saliency}, we propose incorporating saliency guidance for augmentation. Specifically, given an arbitrary normal sample ${\mathbf{x}} \in {\mathcal{X}}$ with a height and width of $H$ and $W$ (${\mathbf{x}} \in \mathbb{R}^{H \times W \times 3}$), respectively, we employ the saliency extractor LayerCAM \cite{jiang2021layercam} to produce a series of salicecy maps $\mathcal{M}=\{\mathbf{m}_1, \cdots, \mathbf{m}_n, \cdots,\mathbf{m}_N \},  \mathbf{m}_n \in \mathbb{R}^{H \times W}$. Benefiting from the design of LayerCAM, the maps in $\mathcal{M}$ represent different levels of attention granularity of the interested object area. Fig. \ref{fig:2} illustrates the overview of our method. As can be observed in the top row, the saliency maps near the output layer (e.g., the map of Layer$\#30$ in  Fig. \ref{fig:2}) characterize the saliency in a more global manner, whereas the maps close to the input layer (e.g., the map of Layer$\#4$ in  Fig. \ref{fig:2}) tend to be more fine-grained. Compared to other saliency-based augmenting strategies \cite{bozorgtabar2023attention,bozorgtabar2022anomaly} which only use a single map, we argue that the saliency map set $\mathcal{M}$ should be fully utilized such that different levels of attention can be reflected in augmentation. We next need to explore how to exploit $\mathcal{M}$ for augmentation.

\begin{algorithm}[t]
    \begin{algorithmic}[1]
        \Require{an arbitrary normal sample ${\mathbf{x}} \in {\mathcal{X}}$}
        \Ensure{$N^s$ paired positive-negative samples}

                   \State Pre-processing: (1) Obtain the positive sample $\widetilde{\mathbf{x}}^+$ by applying weak augmentation to $\mathbf{x}$; (2) Obtain $N^s$ saliency maps $\mathcal{M}^s=\{\mathbf{m}_1, \cdots,  \mathbf{m}_{N^s}\}$ via LayerCAM.
                          
                    \For{$n = 1$ to  $N^s$}
                        
                        \State Apply $K$-means to $\mathbf{m}_n$ to obtain the centroid set $\{c_1, \cdots, c_K\}$

                         \State Randomly sample anchor point pair $(a_1, a_2)$ from the pixels within the cluster with largest centroid intensity $c_k$

                         \State Generate the patch pair $(P_1, P_2)$ on $\widetilde{\mathbf{x}}^+$  with $(a_1, a_2)$

                         \State Obtain the negative sample $\widetilde{\mathbf{x}}_n^{-}$ by swapping the locations of $P_1$ and $P_2$ on $\widetilde{\mathbf{x}}_n^{+}$    
                    
                    \EndFor
                    \State  \textbf{end for}
                    \State Obtain $N^s$ negative samples $\{\widetilde{\mathbf{x}}^{-}_{1}, \cdots, \widetilde{\mathbf{x}}^{-}_{N^s}\}$ 
         
                \State \textbf{return} $N^s$ paired samples $\{(\widetilde{\mathbf{x}}^{+}, \widetilde{\mathbf{x}}^{-}_{1}), \cdots, (\widetilde{\mathbf{x}}^{+}, \widetilde{\mathbf{x}}^{-}_{N^s})\}$
                  
    \end{algorithmic}
    \caption{Sample-wise CutSwap augmentation}
    \label{alg:alg1}
\end{algorithm}

\noindent	\textbf{Saliency Map Clustering.} 
Our general augmentation draws inspiration from CutPaste which requires the selection of two patches. Therefore, two anchor pixels must be determined to produce a patch pair. A straightforward approach is simply adopting pixels with the largest and second-largest saliency intensities for patch generation. However, doing so can noticeably hinder anomaly diversity and cause almost all anomalous regions to be similar. To show the awareness of diversity with respect to saliency intensity, we propose clustering on saliency maps to recognize different saliency modes in an unsupervised fashion. In particular, we employ $K$-means \cite{hartigan1979algorithm} as the clustering approach considering its simplicity and efficiency. Specifically, for the $n$-th saliency map $\mathbf{m}_n$, we directly perform $K$-means to obtain $K$ cluster centroids $\{c_1, \cdots, c_K\}$. Each centroid denotes a pixel in the saliency map with a certain intensity level. Following the saliency guidance, we first select the cluster $c_k$ with the largest intensity value and discard the others. In general, the pixels assigned to this cluster should exhibit the highest saliency on average. We then randomly sample two points ($a_1, a_2$) from this cluster as anchors. Such anchor pairs presents similar semantics with sufficiently significant saliency intensity. We next discuss how to leverage these points for negative sample generation.

\noindent	\textbf{Patch Swapping.} 
Given the anchor pair ($a_1, a_2$), we generate the patch pair ($P_1, P_2$) on the corresponding positive sample $\widetilde{\mathbf{x}}^{+}$ by considering the anchor pixels as the upper-left patch vertices. Here, $P_1$ and $P_2$ have the same size. One naive strategy is to employ the idea of CutPaste by pasting $P_1$ to $P_2$ for the negative sample generation. Instead, we propose swapping the locations of $P_1$ and $P_2$ to yield a negative sample $\widetilde{\mathbf{x}}^{-}$. The primary reason is that the pasting operation generates a negative sample with two identical areas, which is less plausible in real-world scenarios. By contrast, since the patch pair obtained under saliency guidance shows dense semantic correlation, swapping them can impose more subtle yet realistic local irregularities. 

Because we enforce a multilevel saliency mechanism, CutSwap can be performed on every saliency map to eventually yield multiple negative samples. To mitigate the computational overhead, we only apply CutSwap to the saliency map subset $\mathcal{M}^s \subseteq \mathcal{M}$ which includes $N^s < N$ maps. $N^s$ is a pre-determined hyperparameter. Consequently, an arbitrary positive sample $\widetilde{\mathbf{x}}^{+}$ results in a negative sample set $\{\widetilde{\mathbf{x}}^{-}_{1}, \cdots, \widetilde{\mathbf{x}}^{-}_{N^s}\}$. Hence, $N^s$ sample pairs $\{(\widetilde{\mathbf{x}}^{+}, \widetilde{\mathbf{x}}^{-}_{1}), \cdots, (\widetilde{\mathbf{x}}^{+}, \widetilde{\mathbf{x}}^{-}_{N^s})\}$ for one normal sample are prepared. The detailed CutSwap pipeline is described in Alg. \ref{alg:alg1}.

 \begin{figure*}[h]
 \centering
 \includegraphics[width=1\textwidth]{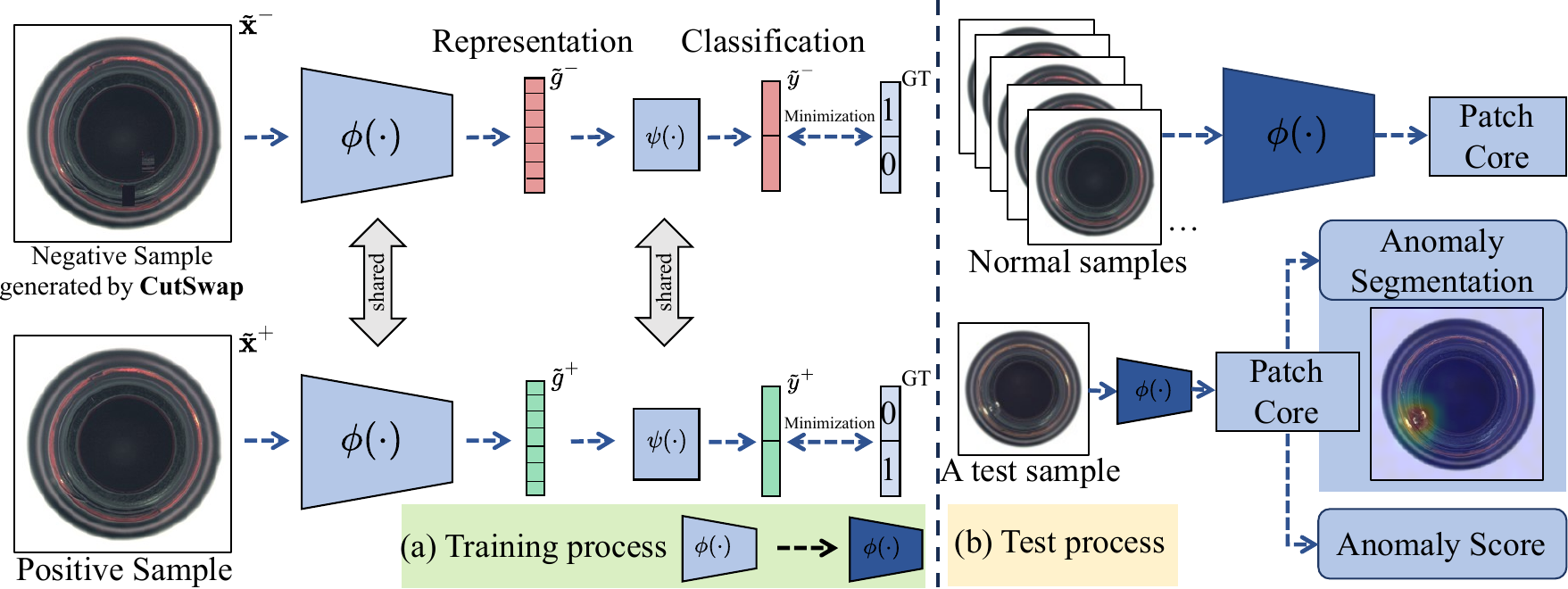}
 \caption{\textbf{Overview of our self-supervised framework.} Training and testing phases are displayed in (a) and (b), respectively. }
 \label{fig:3}
 \end{figure*}

\subsection{Self-supervised AD framework}
\label{sec:self-supervised learning framework}
Up to this point, we have introduced how to obtain the positive-negative sample pair via CutSwap. In the following, we explain the self-supervised learning framework to train the feature extraction network for AD. 

\noindent	\textbf{Training.} Following \cite{li2021cutpaste,zou2022spot}, we tackle  AD by training a powerful feature extraction network $\phi(\cdot)$ such that the latent representations for normal and anomalous samples differ significantly. Specifically, for an arbitrary training sample pair\footnote{Here we omit the index for brevity without the loss of generality.} $(\widetilde{\mathbf{x}}^{+}, \widetilde{\mathbf{x}}^{-})$, we leverage a shared $\phi$ for encoding: 
\begin{equation}
\centering
\label{eq:eq1}
\widetilde{g}^+ =\phi(\widetilde{\mathbf{x}}^+),  
\end{equation}

\begin{equation}
\centering
\label{eq:eq2}
\widetilde{g}^- =\phi(\widetilde{\mathbf{x}}^-),  
\end{equation}
where $\widetilde{g}^{\cdot} \in \mathbf{R}^D$ denotes a $D$-dimensional feature vector. The paired representations $(\widetilde{g}^+,\widetilde{g}^-)$ are then fed into a shared multilayer perceptron (MLP) $\psi$ to predict the label pair, respectively:

\begin{equation}
\centering
\label{eq:eq3}
\widetilde{y}^+ =\psi(\widetilde{g}^+),
\end{equation}

\begin{equation}
\centering
\label{eq:eq4}
\widetilde{y}^- = \psi(\widetilde{g}^-).   
\end{equation}
$\widetilde{y}^{\cdot} \in \mathbb{R}^2$ refers to the classification result.

To encourage the feature extractor $\phi(\cdot)$ to effectively encode distinct representations, we enforce the self-supervised learning policy by designing the loss as a standard binary cross-entropy. For the whole set of normal samples, our loss is given by 
\begin{equation}
\centering
\label{eq:eq5}
\mathcal{L}_{CS}=\mathbb{E}_{x\in \mathcal{X}}\left \{\mathbb{CE}(\widetilde{y}^+,0)+\mathbb{CE}(\widetilde{y}^-,1)   \right \}, 
\end{equation}  
where $\mathbb{CE}(\cdot,\cdot)$ measures the cross-entropy. 

\noindent	\textbf{Testing.} For an arbitrary unseen sample, the testing phase addresses two problems: anomaly score evaluation and segmentation. We realize this by adopting Patch Core \cite{roth2022towards} (PC). In particular, we first train the PC using the original samples with our trained feature extractor $\phi$
to obtain a memory bank that broadly stores the patch-based normal image features. During testing, the PC measures the distance between the patch-level test feature and the collected feature in the memory bank for further nearest neighbor calculation. As shown in Fig. \ref{fig:3}(b), PC produces the final anomaly score and anomaly segmentation results.



\section{Experiment}

\label{Experiment}

In this section, we report experimental results against other AD methods to evaluate the effectiveness of our method quantitatively and qualitatively. We also conduct ablation studies for a detailed analysis.



\noindent \textbf{Dataset.} To be consistent with previous 2D AD literatures \cite{bergmann2020uninformed, zou2022spot}, we evaluate our method on the MVTec AD \cite{bergmann2021mvtec} and VisA \cite{zou2022spot} datasets. \par
\begin{itemize}
\item \textbf{MVTec AD} is a widely used large-scale benchmark dataset for industrial AD tasks. It consists of 15 categories of industrial products (e.g., tiles, bottles) with 5,354 high-resolution color images. 73 different types of anomalies are included. The training set involves only the anomaly-free samples, while the test set covers both anomalous and anomaly-free unseen samples.

\item \textbf {Visual Anomaly (VisA)} is a large-scale industrial dataset for both 1- and 2-class anomaly detection, which contains 10,821 high-resolution color images with 9,621 normal and 1,200 anomalous samples. There are 12 object types that are further categorized into complex structures, multiple instances, and single instances to satisfy different detection tasks.
\end{itemize}

\noindent \textbf{Implementation details.} We adopt the ResNet18 \cite{he2016deep} architecture to construct our feature extractor $\phi$. As suggested in \cite{he2016deep}, we first pre-train it on ImageNet \cite{deng2009imagenet} dataset and then fine-tune it on MVTec AD and VisA for 256 epochs using the SGD optimizer with an initial learning rate of 0.03, respectively. Note that the fine-tuning on MVTec AD and VisA is individually performed when experimenting on each dataset. We set ($N, N^s, K, D$) to (30, 5, 4, 512). More precisely, each original image data will lead to the generation of five positive-negative training sample pairs. 
To produce the saliency map subset $\mathcal{M}^s$, we empirically select the layers indexed as 4, 9, 16, 23, and 30, considering the trade-off between performance and computational overhead. 

In our implementation, we follow the 3-way training manner introduced in \cite{li2021cutpaste} by further preparing a scarred CutSwap negative sample set $\mathcal{X}^{-}_{S}$. In particular, during patch generation for samples in $\mathcal{X}^{-}_{S}$, we control the patch sizes to create long-thin rectangular boxes to mimic scars intentionally. After patch swapping, these ``scars'' are eventually rotated to increase anomaly variations. Given the sample triplet in ($\mathcal{X}^+, \mathcal{X}^-, \mathcal{X}^-_{S}$), we perform multi-class classification by extending the binary cross-entropy loss in Eq. \ref{eq:eq5} to the multi-class version to train the feature extractor $\phi$. This is to leverage the inherent attribute difference of anomaly types in $\mathcal{X}^{-}$ and $\mathcal{X}^{-}_{S}$
to better empower $\phi$.


\begin{table*}[]
\centering
    \caption{\centering \textbf{Quantitative evaluation of AD performance on MVTec AD with Patch Core (PC) coreset set to 0.001.} The best and second-best results are indicated in bold and underlined, respectively.}
\label{tab:tab1}

\begin{tabular}{ccccccc}
\toprule
\multirow{2}{*}{Category} & \multicolumn{2}{c}{CutPaste + PC} & \multicolumn{2}{c}{ResNet18 + PC} & \multicolumn{2}{c}{CutSwap + PC} \\ \cline{2-7} 
                          & Image-level   & Pixel-level  & Image-level   & Pixel-level  & Image-level  & Pixel-level  \\ \hline
Bottle                    & \textbf{1.000} & 0.952          & \textbf{1.000}         & \underline{0.972}         & \textbf{1.000}  & \textbf{0.975}        \\
Cable                     & 0.934         & 0.936          & \underline{0.945}         & \underline{0.948}        & \textbf{0.957}        & \textbf{0.949}        \\
Capsule                   & 0.812         & 0.895          & \textbf{0.957}         & \underline{0.969}        & \underline{0.945}        & \textbf{0.970}         \\
Carpet                    & 0.846         & 0.976          & \textbf{0.987}         & \underline{0.986}        & \underline{0.975}        & \textbf{0.988}         \\
Grid                      & \underline{0.991}       & \underline{0.938}          & 0.931         & 0.925         & \textbf{0.993}        & \textbf{0.940}         \\
Hazelnut                  & \underline{0.990}         & 0.973          & \textbf{1.000}        & \textbf{0.978}         & \textbf{1.000}        & \underline{0.977}         \\
Leather                   & \textbf{1.000}  & \textbf{0.989}          & \textbf{1.000}         & \underline{0.988}        & \textbf{1.000}        & 0.987         \\
Metal nut                & 0.918         & 0.938          & \underline{0.951}        & \underline{0.949}         & \textbf{0.972}        & \textbf{0.957}         \\
Pill                      & 0.847         & 0.866          & \underline{0.895}         & \textbf{0.949}         & \textbf{0.940}        & \underline{0.928}         \\
Screw                     & \underline{0.696}        & \textbf{0.867}          & 0.675         & 0.807         & \textbf{0.732}        & \underline{0.832}         \\
Tile                      & \textbf{0.999}         & \underline{0.920}          & 0.982         & \textbf{0.921}        & \underline{0.983}   & 0.913         \\
Toothbrush                & \textbf{0.953}         & 0.939          & 0.914         & \underline{0.960}         & \underline{0.917}    & \textbf{0.977}       \\
Transistor                & 0.938         & 0.914          & \textbf{0.976}         & \underline{0.927}         & \underline{0.944}        & \textbf{0.929}        \\
Wood                      & \underline{0.994}         & \textbf{0.932}         & 0.990         & 0.919         & \textbf{0.999}      & \underline{0.927}         \\
Zipper                    & \textbf{0.998}         & \textbf{0.980}         & \underline{0.975}        & \underline{0.975}         & \underline{0.975}       & 0.960          \\ \hline
\rule{0pt}{8pt}
Average          & 0.928         & 0.934          & \underline{0.945}        & \underline{0.945}         & \textbf{0.955} & \textbf{0.947}        \\ \bottomrule

\end{tabular}
\end{table*}

\begin{table*}[]
\centering
\caption{\centering \textbf{Quantitative evaluation of AD performance on MVTec AD with Patch Core (PC) coreset set to 0.01.} The best and second-best results are indicated in bold and underlined, respectively.}
\label{tab:tab2}
\scalebox{0.9}{

\begin{tabular}{ccccccccccc}
\toprule
\multirow{2}{*}{Category} & \multicolumn{2}{c}{CutPaste + PC} & \multicolumn{2}{c}{ResNet18 + PC} & \multicolumn{2}{c}{STPM \cite{wang2021student_teacher}} & \multicolumn{2}{c}{InTra \cite{pirnay2022inpainting}} & \multicolumn{2}{c}{CutSwap + PC} \\ \cline{2-11} 
                          & Image-level   & Pixel-level  & Image-level    & Pixel-level & Image-level  & Pixel-level     & Image-level  & Pixel-level       & Image-level  & Pixel-level               \\ \hline
Bottle                    & \textbf{1.000} & 0.957        & \textbf{1.000} & 0.975           &-            &\textbf{0.988}        &\textbf{1.000} &0.971            & \underline{0.999}   & \underline{0.979}       \\
Cable                     & 0.981         & 0.968         & \underline{0.986} & \underline{0.976}           &-          &0.955        &0.703         &0.910                    & \textbf{0.995}     & \textbf{0.977}   \\
Capsule                   & 0.916         & 0.960         & \textbf{0.956}          & \textbf{0.987}           &-            &0.983        &0.865         &0.977     & \underline{0.947}     & \underline{0.984} \\
Carpet                    & 0.861         & 0.980         & \underline{0.987}         & 0.987           &-            &0.988        & \textbf{0.988}        & \textbf{0.992}    & 0.980      & \underline{0.990}   \\
Grid                      & \underline{0.994}         & 0.951         & 0.949          & 0.961           &-            &\textbf{0.990}        &\textbf{1.000}        &\underline{0.988}       & 0.988  & 0.948      \\
Hazelnut      & \textbf{1.000}        & 0.976         & \textbf{1.000}          & 0.982          &-            &\textbf{0.985}        &0.957         &\underline{0.983}      & \underline{0.998}   & 0.980   \\
Leather                   & \textbf{1.000}         & 0.989         & \textbf{1.000}         & 0.989       &-         &\underline{0.993}       &\textbf{1.000}         &\textbf{0.995}                    & \textbf{1.000}         & 0.986      \\
Metal nut                & 0.951         & 0.967         & \underline{0.991}         & \underline{0.976}           &-            &\underline{0.976}        &0.969         &0.933            & \textbf{0.998}         & \textbf{0.978}              \\
Pill                      & 0.895         & 0.913         & \underline{0.948}          & 0.970           &-            &\underline{0.978}       &0.902         &\textbf{0.983}               & \textbf{0.960}         & 0.962              \\
Screw                     & 0.848         & 0.970         & \underline{0.949}         & \underline{0.985}          &-            &0.983        &\textbf{0.957}         &\textbf{0.995}       & 0.871         & 0.975              \\
Tile                      & \textbf{0.998}         & 0.917         & \underline{0.987}         & 0.930           &-            &\textbf{0.974}        &0.982         &\underline{0.944}       & \underline{0.987}         & 0.914              \\
Toothbrush                & 0.983         & 0.973         & \underline{0.989}        & 0.986   &-            &\textbf{0.989}   &\textbf{1.000}         &\textbf{0.989}                    & \textbf{1.000}         & \underline{0.988}              \\
Transistor                & 0.980         & 0.954         & \textbf{0.999}          & \underline{0.965}           &-            &0.825        &0.958         &0.961                    & \underline{0.998}         & \textbf{0.970}              \\
Wood                      & \underline{0.995}         & \underline{0.940}         & 0.986          & 0.922           &-            &\textbf{0.972}        &0.975         &0.887                    & \textbf{0.997}         & 0.937              \\
Zipper                    & \textbf{0.998}        & 0.980         & 0.983          & \underline{0.985}          &-            &\underline{0.985}        &\underline{0.994}        &\textbf{0.992}         & 0.982         & 0.971      \\ \hline 
\rule{0pt}{8pt}
Average                   & 0.960         & 0.959         & \textbf{0.981} & \textbf{0.972}  &0.955        &\underline{0.970}       &0.950         &0.966                    & \underline{0.980}         & \underline{0.970}                 \\ \bottomrule
\end{tabular}}
\end{table*}

\begin{table*}[]
\caption{\centering \textbf{Quantitative evaluation of AD performance on VisA with Patch Core (PC) coreset set to 0.001.} The best and second-best results are indicated in bold and underlined, respectively.}
\label{tab:tab3}
\centering
\begin{tabular}{ccccccc}
\toprule
\multirow{2}{*}{Category} & \multicolumn{2}{c}{CutPaste + PC} & \multicolumn{2}{c}{ResNet18 + PC} & \multicolumn{2}{c}{CutSwap + PC} \\ \cline{2-7} 
                          & Image-level   & Pixel-level  & Image-level   & Pixel-level  & Image-level  & Pixel-level  \\ \hline
Candles                   & 0.889          & 0.875         & \textbf{0.970}         & \textbf{0.977}        & \underline{0.926}        & \underline{0.957}   \\
Capsules                   & \textbf{0.755}          & \textbf{0.955}        & 0.706          & 0.930        & \underline{0.741}         & \underline{0.944}         \\
Cashew                    & 0.883          & 0.951         & \textbf{0.917}          & \textbf{0.984}       & \underline{0.887}        & \underline{0.970}       \\
Chewinggum                & \textbf{0.988}          & \underline{0.983}         & \underline{0.975}         & \textbf{0.989}        & 0.972         & \textbf{0.989}         \\
Fryum                     & \textbf{0.943}          & 0.870         & 0.889          & \underline{0.908}       & \underline{0.923}         & \textbf{0.915}       \\
Macaroni1                 & 0.774          & 0.910         & \underline{0.797}         & \underline{0.923}       & \textbf{0.852}         & \textbf{0.951}       \\
Macaroni2                 & \underline{0.696}         & \underline{0.837}         & \textbf{0.748}          & 0.789        & 0.686         & \textbf{0.866}       \\
Pcb1                      & \textbf{0.947}          & \textbf{0.996}         & 0.856          & \underline{0.990}        & \underline{0.941}        & \textbf{0.996}     \\
Pcb2                      & \underline{0.885}          & \underline{0.968}         & \textbf{0.891}          & \textbf{0.972}        & 0.874         & 0.952         \\
Pcb3                      & 0.827          & \underline{0.980}         & \underline{0.860}          & \textbf{0.985}       & \textbf{0.890}         & 0.977         \\
Pcb4                      & 0.975          & 0.970         & \textbf{0.982}         & \underline{0.975}       & \underline{0.981}         & \textbf{0.976}         \\
Pipe fryum                 & 0.763          & 0.961         & \textbf{0.988}          & \textbf{0.986}       & \underline{0.898}        & \underline{0.974}         \\ \hline 
\rule{0pt}{8pt}
Average         & 0.860          & 0.938         &\textbf{0.882}  & \underline{0.951}       & \underline{0.881}        & \textbf{0.956} \\ \bottomrule

\end{tabular}
\end{table*}

\begin{table*}[]
\centering
\caption{\centering \textbf{Quantitative evaluation of AD performance on VisA with Patch Core (PC) coreset set to 0.01.} The best and second best results are in bold and underlined, respectively.}
\label{tab:tab4}
\scalebox{0.9}{

\begin{tabular}{ccccccccccc}
\toprule
\multirow{2}{*}{Category} & \multicolumn{2}{c}{CutPaste + PC} & \multicolumn{2}{c}{ResNet18 + PC} & \multicolumn{2}{c}{SPD + Padim  \cite{zou2022spot}}  & \multicolumn{2}{c}{Draem \cite{zavrtanik2021draem}} & \multicolumn{2}{c}{CutSwap + PC} \\ \cline{2-11} 
                          & Image-level   & Pixel-level  & Image-level   & Pixel-level         & Image-level   & Pixel-level     & Image-level   & Pixel-level   & Image-level  & Pixel-level  \\ \hline
Candles                    & 0.909         & 0.946         & \textbf{0.984}          & \textbf{0.986}           & 0.891          & 0.973     &0.823         &0.870    & \underline{0.956}        & \underline{0.983}   \\
Capsules                   & 0.704         & \textbf{0.974}         & 0.718          & \underline{0.965}             & 0.681          & 0.863     &\textbf{0.773}   &0.937      & \underline{0.732}      & \textbf{0.974}   \\
Cashew                    & 0.919         & 0.971         & \underline{0.935}          & \textbf{0.989}             & 0.905          & 0.861     &\textbf{0.942}         &0.947    & 0.904         & \underline{0.985}   \\
Chewinggum                & 0.990         & \underline{0.988}        & 0.983          & \textbf{0.990}             & \underline{0.993}         & 0.969     &0.934     &0.975      & \textbf{0.994}     & \textbf{0.990}    \\
Fryum                     & 0.932         & 0.929         & 0.923          & \underline{0.949}            & 0.898          & 0.880     &\textbf{1.000}    &\textbf{0.975}        & \underline{0.941}         & 0.938    \\
Macaroni1                 & 0.831         & 0.944         & \textbf{0.890}         & \underline{0.983}   & 0.857          & \textbf{0.988}     &0.703         &0.958     & \underline{0.874}     & \underline{0.983}    \\
Macaroni2                 & 0.712         & 0.953         & \textbf{0.756}         & 0.952             & 0.708          & \underline{0.960}     &0.713         &0.941       & \underline{0.722}        & \textbf{0.964}    \\
Pcb1                      & \textbf{0.959}  & \textbf{0.998}  & 0.939  & \textbf{0.998}    & 0.927     & 0.977     &0.713         &\underline{0.986}          & \underline{0.956}        & \textbf{0.998}   \\
Pcb2                      & \textbf{0.944}      & \underline{0.978}    & \underline{0.930}       & \textbf{0.979}     & 0.879          & 0.972     &0.897         &0.925                  & 0.924         & 0.976     \\
Pcb3                      & \underline{0.929}         & \underline{0.987}   & \textbf{0.932}          & \textbf{0.988}       & 0.854          & 0.967     &0.731         & 0.938     & 0.927     & \underline{0.987} \\
Pcb4                      & 0.986         & \underline{0.984}  & \underline{0.991}      & 0.982       & \underline{0.991}     & 0.892     & 0.913        &0.958        & \textbf{0.996}        & \textbf{0.985}    \\
Pipe fryum                 & 0.851         & 0.977         & \textbf{0.996}         & \textbf{0.989}          & \underline{0.956}          & 0.954    &0.941         &0.818        & 0.926     & \underline{0.984}     \\ \hline  
\rule{0pt}{8pt}
Average                   & 0.889 & \underline{0.969}  & \textbf{0.915} & \textbf{0.979}    & 0.878           &0.938      &0.841         &0.888                      & \underline{0.904} & \textbf{0.979} \\ \bottomrule

\end{tabular}}
\end{table*}

 \begin{figure*}[h]
 \centering
 \includegraphics[width=1\textwidth]{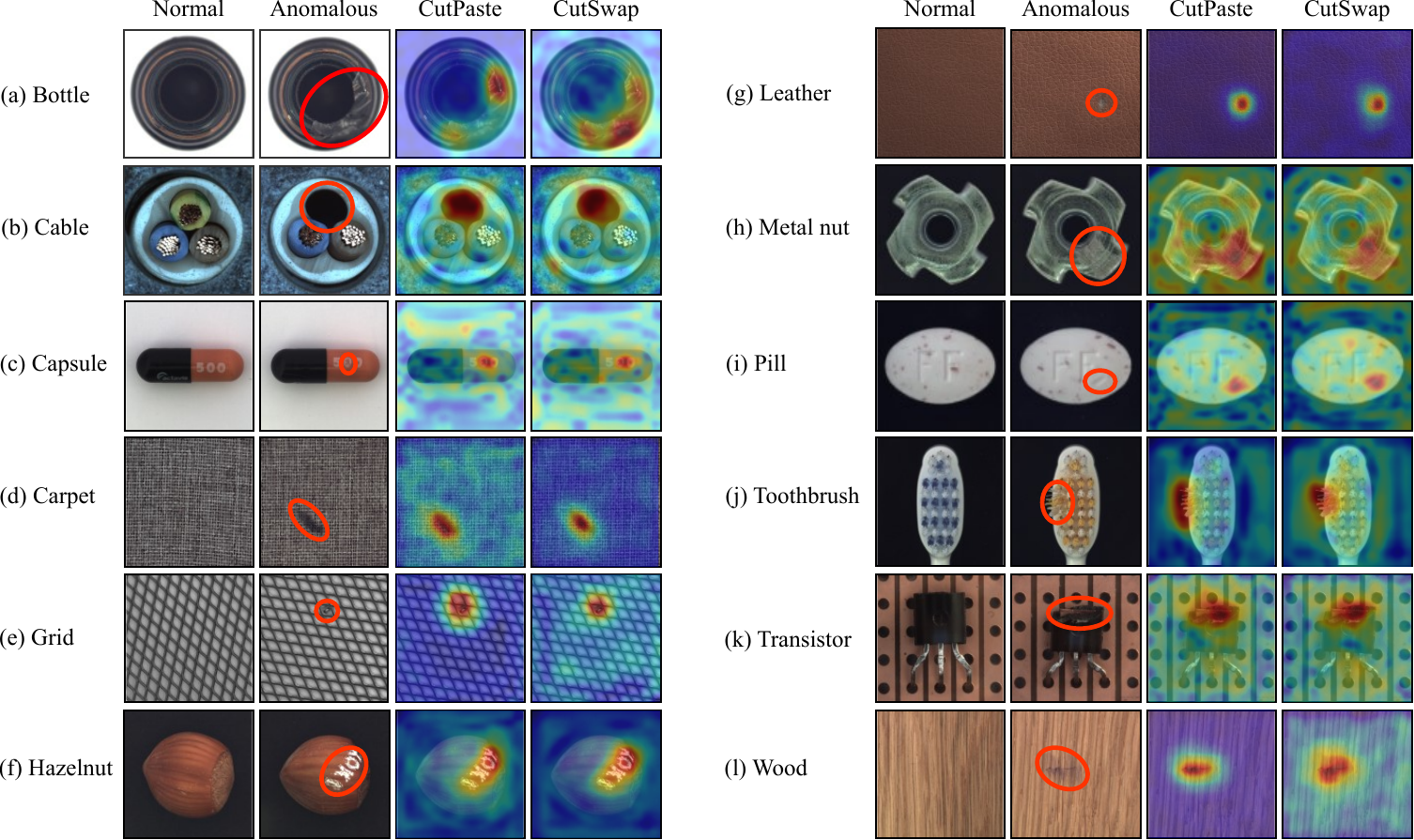}
     \caption{\textbf{Qualitative pixel-wise detection results on MVTec AD.} Circled areas in red denote the ground-truth anomalous regions. }
 \label{fig:4}
 \end{figure*}

  \begin{figure*}[h]
 \centering
 \includegraphics[width=1\textwidth]{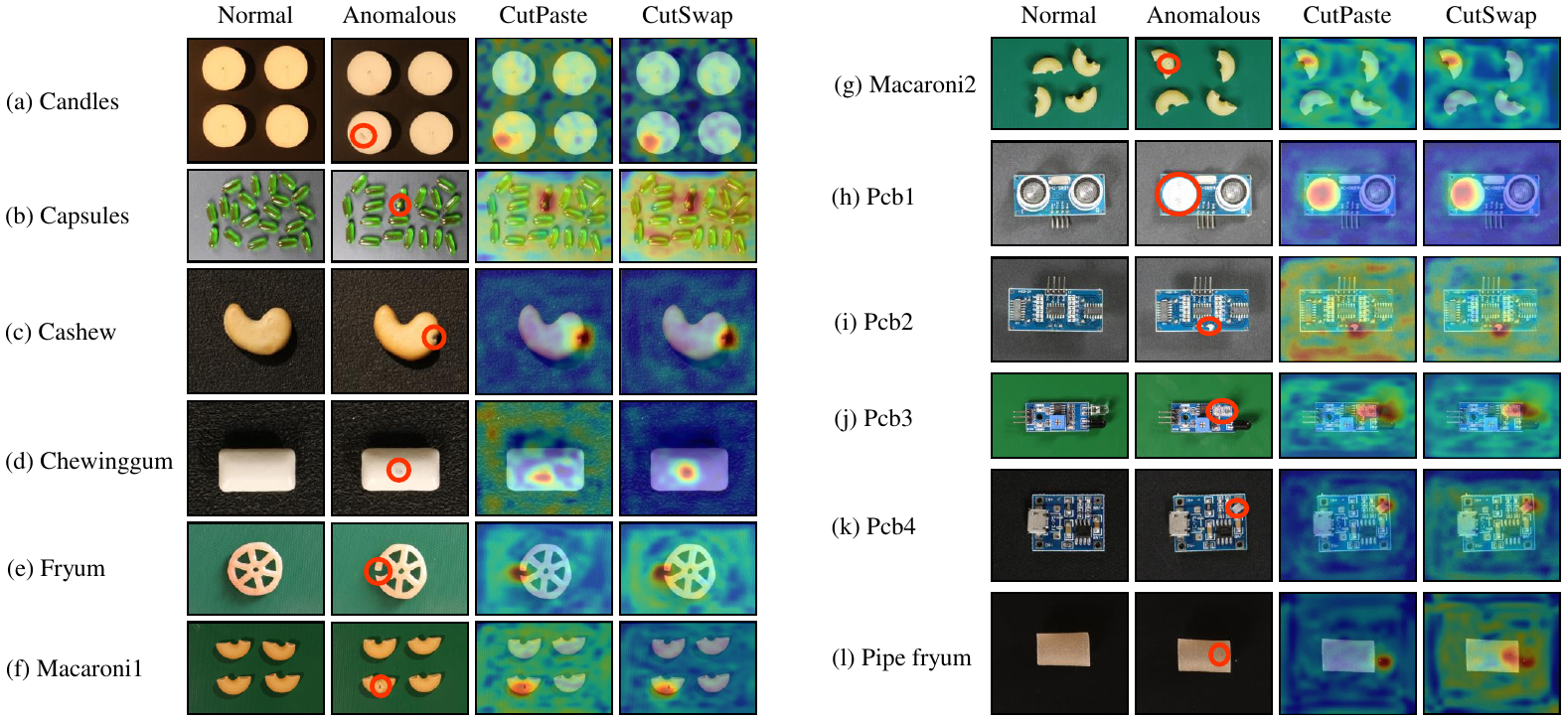}
 \caption{\textbf{Qualitative pixel-wise detection results on VisA.} Circled areas in red denote the ground-truth anomalous regions. }
 \label{fig:5}
 \end{figure*}

\noindent	\textbf{Evaluation Metric.}
Following \cite{roth2022towards,zou2022spot}, we use the area under the receiver operating characteristic curve (AUC) to evaluate our method. AUC calculates the entire area under the Receiver Operating Characteristic (ROC) curve for assessment. A higher AUC score indicates a more powerful detection capacity. We report both image- and pixel-level AUCs for evaluation.


\subsection{Evaluation}

\noindent	\textbf{Quantitative Results.}
First, we report the quantitative detection results against existing AD methods. \textit{Note that, since our goal is to analyze the effectiveness of CutSwap augmentation rather than the strength of complex feature extraction architectures, we generally limit our discussion on lightweight models for fair comparisons.} In particular, for MVTec AD, we compare against CutPaste \cite{li2021cutpaste}, STPM \cite{wang2021student_teacher}, and InTra \cite{pirnay2022inpainting}, while for VisA, we compare with CutPaste\cite{li2021cutpaste}, SPD \cite{zou2022spot}, and Draem \cite{zavrtanik2021draem}.  CutPaste and SPD are two self-supervised AD methods. STPM and CutPaste use ResNet18, while SPD adopts ResNet50. InTra and Draem are two reconstruction-based unsupervised AD approaches. We also compare against the original pre-trained ResNet18 architecture without self-supervised tuning to gain more insights.

The category-wise AD results are summarized in Tab. \ref{tab:tab1} $\sim$ \ref{tab:tab4}. For each dataset, we report the results under two Patch Core (PC) coreset settings. A greater coreset value indicates a larger coverage of patch feature types. It can be observed that under a smaller coreset setting, where only limited features can be utilized, our method achieves better AD results than CutPaste or untuned ResNet18 on both datasets.  This is because benefiting from the multilevel strategy and the swap operation that jointly contribute to semantics preservation, our network is endowed with more powerful capacity to capture key features. This well demonstrates the low sensitivity of our feature extractor to available feature quantity within Patch Core.
For both unsupervised methods STPM and InTra, and the self-supervised method CutPaste on MVTec AD, we can observe from Tab. \ref{tab:tab2} that CutSwap consistently achieves higher detection accuracy. Specifically, although CutPaste also performs self-supervised learning, since it does not show awareness of semantics, the feature extraction network can be restricted to encode powerful latent presentations. Similar results can be observed on the self-supervised method SPD (Tab. \ref{tab:tab4}) on the VisA dataset. In addition, unsupervised methods (i.e., InTra and Draem) perform less satisfactory because no augmentation is involved. From the above analysis, we can verify that due to the sample superiority, training with the CutSwap sample pair yields state-of-the-art quantitative AD results. \par

\begin{table}[tb]
\caption{\centering \textbf{Ablative evaluation of cluster number setting.} Patch Core coreset is set to 0.001. $K$=1 refers to randomly sampling anchor points on the entire image for swapping.}
\label{tab:tab5}
\centering
\begin{tabular}{ccc}
\toprule
Value of $K$ & Image-level AUC & Pixel-level AUC \\ \hline
1           & 0.936                 & \textbf{0.952}               \\
2            & 0.937                 & 0.950               \\
3            & 0.938                 & 0.945                 \\
4            & \textbf{0.955}        & 0.947               \\
5            & 0.940                 & 0.939                 \\
6            & 0.927                 & 0.934                 \\ \bottomrule
\end{tabular}
\end{table}

\begin{table}[tb]
\caption{\centering \textbf{Ablative evaluation of cluster selection on via average AUC.} Patch Core coreset is set to 0.001. }
\label{tab:tab6}
\centering
\begin{tabular}{ccc}
\toprule
Cluster        & Image-level AUC & Pixel-level AUC \\ \hline
Random cluster & 0.931        &\textbf{0.947}        \\
Min cluster    & 0.937        & 0.940        \\
Max cluster    & \textbf{0.955} & \textbf{0.947}       \\ \bottomrule
\end{tabular}
\end{table}

\begin{table*}[]
\caption{\centering \textbf{Ablative evaluation of saliency map combinations via average AUC.} Patch Core coreset is set to 0.001. }
\label{tab:tab7}
\centering
\begin{tabular}{cccccccc}
\toprule
\multirow{2}{*}{Layer combination} & \#4     & \#16    & \#30    & \#4 + \#16 & \#16 + \#30 & \#4 + \#16 + \#30 & \#4 + \#9 + \#16 + \#23 + \#30 \\ \cline{2-8} 
                                & $N^s$=1 & $N^s$=1 & $N^s$=1 & $N^s$=2    & $N^s$=2     & $N^s$=3           & $N^s$=5                       \\ \hline
Image-level AUC                 & 0.940   & 0.939       & 0.936   & 0.939      & 0.949       & 0.939             & \textbf{0.955}                        \\ 
Pixel-level AUC                 & 0.946   & 0.950       & 0.946   & \textbf{0.951}      & 0.950       & 0.950             & 0.947                         \\ \bottomrule
\end{tabular}
\end{table*}

\begin{table*}[tb]
\caption{\centering \textbf{Ablative evaluation of anchor point selection strategies via average AUC.} }
\label{tab:tab8}
\centering
\begin{tabular}{lcccc}
\toprule
\multicolumn{1}{c}{\multirow{2}{*}{Strategy}} & \multicolumn{2}{c}{PC Coreset=0.001}                                               & \multicolumn{2}{c}{PC Coreset=0.01}                                                \\ \cline{2-5} 
\multicolumn{1}{c}{}                          & \multicolumn{1}{l}{Image-level AUC} & \multicolumn{1}{l}{Pixel-level AUC} & \multicolumn{1}{l}{Image-level AUC} & \multicolumn{1}{l}{Pixel-level AUC} \\ \hline
Saliency sorting                                 & 0.947                                  & 0.940                                  & 0.973                                  & 0.966                                  \\
$K$-means                                       & \textbf{0.955}                        & \textbf{0.947}                         & \textbf{0.980}                       & \textbf{0.970}                                 \\ \bottomrule
\end{tabular}
\end{table*}

\noindent	\textbf{Qualitative Results.}
 We qualitatively assess our method by visualizing the pixel-wise AD results on MVTec AD and VisA against CutPaste \cite{li2021cutpaste} in Fig. \ref{fig:4} and Fig. \ref{fig:5}, respectively. We notice that despite the fact that CutPaste generally yields visual detection results comparable to ours, it tends to induce large false-positive regions on the anomaly segmentation maps. For example, for the categories of the grid (Fig. \ref{fig:4}(e)), metal nut (Fig. \ref{fig:4}(h)), and toothbrush (Fig. \ref{fig:4}(j)) whose anomaly regions are small, CutPaste may wrongly treat the neighboring pixels as anomalous regions during detection, which leads to false-positives. Also, the detection results of candles (Fig. \ref{fig:5}(a)) and macaroni1 (Fig. \ref{fig:5}(f)) on the VisA dataset display a similar issue. This is also reflected in the low pixel-level AUC scores in Tab. \ref{tab:tab1} and Tab. \ref{tab:tab3}. In contrast to this, because CutSwap learns from the negative samples via semantic-aware augmentation guided by saliency, it handles the subtle anomalies better than CutPaste.  

 Furthermore, in addition to the false-positive issue, CutPaste sometimes fails to localize anomalies. This can be observed in both anomalies with large areas on MVTec AD (bottle, Fig. \ref{fig:4}(a)), and subtle anomalies on VisA (pipe fryum, Fig. \ref{fig:5}(l)). We suspect that the failure stems from the invalid samples introduced in the CutPaste augmentation mechanism, which hinders the networks from learning quality feature embeddings. From the above analysis, we can qualitatively confirm that our CutSwap generates satisfactory pixel-wise AD results, especially for tiny yet challenging cases, compared with the prior competitor CutPaste.

\subsection{Ablation studies}
To provide more understanding of our method, we conduct ablation studies for detailed analysis from the following aspects:  \par
\noindent	\textbf{Cluster Number $K$.}
First, we investigate the influence of the cluster number setting in applying $K$-means. Tab. \ref{tab:tab5} reports the quantitative results on MVTec AD by varying $K$ during training. We notice that an overly large or small selection of $K$ degrades performance. Intuitively, a smaller $K$ would cause the target cluster for anchor selection to contain unexpected pixels with a low saliency score, whereas a greater $K$ can cause the size of the target cluster to be too small to ensure anomaly diversity. Thus, we empirically set $K$ to 4 in all our experiments, considering the overall performance trade-off between the image- and pixel-level AUCs.

\noindent	\textbf{Target Cluster Selection.} Our key motivation is to let the cluster with the maximum saliency score guide the negative sample generation. To examine the validity of this, we switch the cluster for producing anchors with the centroids of lower saliency for comparison. Specifically, we experiment under two selection scenarios: random cluster and minimum-saliency cluster, and summarize the results in  Tab. \ref{tab:tab6}. We can see that selecting the maximum cluster in the CutSwap configuration contributes the best to the detection performance, further evidencing the significance of introducing saliency guidance.

\noindent	\textbf{Multilevel Saliency Guidance.} Our method resorts to multilevel saliency maps to pursue swapping guidance. We systematically investigate the impact of layer combinations used in augmentation and report the results in Tab. \ref{tab:tab7}. As can be seen in Tab. \ref{tab:tab7}, using multilevel saliency guidance generally leads to improved performance compared to single-scale guidance. 
This is because the saliency maps yield different levels of attention to locate the object, thus providing more clues for augmentation. Furthermore, for Image-level AUC, 
leveraging all saliency maps (i.e., $N^s=5$) achieves the best result, while for the pixel-level case, the AUC change tends to be steady. We assume the reason is that a coarse-to-fine combination of attention granularity empowers the feature extractor with more global cues for detection. Based on the analysis, we can verify that multi-scale saliency guidance plays a positive role in CutSwap augmentation, and setting $N^s$ to 5 ensures better augmentation quality.


\noindent	\textbf{Validity of Clustering.}
Our anchor points for swapping are based on the clustering algorithm to exploit the high-saliency region guidance. Straightforwardly, one can simply select the pixels with the highest saliency scores for swapping. To validate the significance of the clustering procedure in our strategy, we ablate the $K$-means and enforce a simple counterpart for comparison. Specifically, we first sort the pixels in the anomaly-free image according to the saliency score and then generate anchor pairs from the top 300 pixels for swapping. As reported in Tab. \ref{tab:tab8}, introducing clustering in our selection pipeline outperforms the saliency sorting. This is because, although the pixels within the cluster are generally with high saliency scores, they also provide a part of average-salient pixels readily for swapping, thus encouraging diversity. We can then confirm the significance of using clustering in augmentation.


\section{Conclusion}
In this study, we introduced a novel data augmentation method - CutSwap that generates negative samples by swapping patch pairs on positive ones. We proposed resorting to the multilevel attention mechanism by following saliency guidance to impose stronger semantic clues to anchor pixel selection for swapping. We further incorporated a clustering procedure such that patch swapping can be enforced within high-saliency regions while encouraging anomaly diversity. Our CutSwap results in subtle yet realistic negative samples, better retaining the original semantics. Extensive evaluations and ablative experiments on two large-scale datasets quantitatively and qualitatively demonstrated that the feature extractor trained with CutSwap samples contributes to state-of-the-art performance on the image AD task.

Although CutSwap shows satisfactory accuracy for AD, it depends heavily on the quality of the saliency extraction method. Once the saliency map induces an undesired attention distribution, the augmented samples are prone to be less natural. This could be resolved by weighting and fusing these attention maps for global saliency refinement to better guide augmentation. We would like to explore this interesting topic in the future.



\section*{Acknowledgment}
This work is partially supported by JSPS KAKENHI Grant Number JP23K10712. The China Scholarship Council (CSC) funded the first author under Grant Number 202108330097.

\bibliographystyle{elsarticle-num}
\bibliography{refs}

\end{document}